\documentclass{article}
\usepackage[utf8]{inputenc}
\usepackage[T1]{fontenc}
\usepackage{hyperref}
\usepackage{url}
\usepackage{booktabs}
\usepackage{amsfonts}
\usepackage{amsmath}
\usepackage{amssymb}
\usepackage{nicefrac}
\usepackage{microtype}
\usepackage{graphicx}
\usepackage{xcolor}
\usepackage{multirow}
\usepackage{subcaption}
\usepackage[margin=1in]{geometry}

\title{Forward versus Backward: Comparing Reasoning Objectives in Direct Preference Optimization}

\author{
  \begin{tabular}[t]{c}
    \small Murtaza Nikazad \\
    \small Department of Math and Computer Science\\
    \small Davidson College\\
    \small \texttt{munikzad@davidson.edu}
  \end{tabular}
  \hspace{2em}
  \begin{tabular}[t]{c}
    \small Raghuram Ramanujan \\
    \small Department of Math and Computer Science\\
    \small Davidson College\\
    \small \texttt{raramanujan@davidson.edu}
  \end{tabular}
}

\date{}

\begin{document}

\maketitle

\begin{abstract}
Large language models exhibit impressive reasoning capabilities yet frequently generate plausible but incorrect solutions, a phenomenon commonly termed hallucination. This paper investigates the effect of training objective composition on reasoning reliability through Direct Preference Optimization. Two complementary training signals are examined: forward chain-of-thought generation, which trains the model to produce correct reasoning traces, and backward verification, which trains the model to verify and acknowledge errors in candidate solutions. Experiments on GSM8K reveal a fundamental trade-off between these objectives. Forward-only DPO training achieves the highest accuracy improvement, increasing from 83.1\% to 86.6\% (+3.5 percentage points), while backward-only training yields minimal accuracy gains but substantially reduces the false positive rate from 13.4\% to 4.3\%. Notably, both training variants reduce acknowledgement rate compared to the baseline, suggesting that preference optimization increases model confidence in its outputs. These findings indicate that forward and backward reasoning objectives provide distinct and complementary learning signals: forward training improves problem-solving capability, while backward training improves verification calibration. The complete training and evaluation pipeline, implemented efficiently through Low-Rank Adaptation, is released to facilitate further research.
\end{abstract}

\section{Introduction}

Large language models have demonstrated remarkable capabilities in complex reasoning tasks, achieving strong performance on mathematical problem-solving \cite{cobbe2021gsm8k}, code generation, and scientific reasoning. Chain-of-thought prompting \cite{wei2022chain} has emerged as a powerful technique for eliciting step-by-step reasoning, making the model's problem-solving process more transparent and often more accurate. However, a persistent challenge remains: these models frequently produce confident, coherent reasoning chains that nevertheless lead to incorrect answers \cite{huang2023survey}. More problematically, when presented with their errors, models often fabricate justifications rather than acknowledging mistakes.

This paper investigates a fundamental question: how do different training objectives affect both reasoning accuracy and error recognition capability? Specifically, this work examines whether training models on forward reasoning traces (problem to solution) produces different effects than training on backward verification traces (checking whether a candidate answer is correct).

To study this question, two distinct training configurations are compared within a Direct Preference Optimization framework. Forward-only training optimizes the model to prefer correct reasoning traces over incorrect ones. Backward-only training optimizes the model to prefer accurate verification verdicts, training it to output PASS for correct answers and FAIL for incorrect ones. This experimental design enables isolation of the contribution of each training signal.

The proposed method employs Direct Preference Optimization \cite{rafailov2023direct}, which directly optimizes the policy to prefer chosen completions over rejected ones without requiring a separate reward model. Low-Rank Adaptation \cite{hu2022lora} enables parameter-efficient fine-tuning, making the approach practical for standard GPU hardware.

This paper makes four primary contributions. First, it provides empirical comparison of forward-only and backward-only DPO training for reasoning tasks, revealing distinct effects on accuracy and verification capability. Second, it introduces the acknowledgement rate metric, which measures how often a model correctly identifies its own incorrect answers, as a key indicator of reasoning reliability. Third, it demonstrates that forward DPO training achieves substantial accuracy gains (+3.5 percentage points on GSM8K), while backward training primarily improves verification calibration. Fourth, it releases a complete, reproducible pipeline including data generation with rejection sampling, preference training, and comprehensive evaluation.

\section{Related Work}

\paragraph{Chain-of-Thought Reasoning.}
Chain-of-thought prompting \cite{wei2022chain} significantly improves language model performance on reasoning tasks by eliciting intermediate computational steps. Wang et al. \cite{wang2023selfconsistency} extended this approach through self-consistency, where multiple reasoning paths are sampled and aggregated through majority voting. Yao et al. \cite{yao2023tree} proposed tree-of-thoughts, which structures reasoning as explicit search over a tree of possible reasoning steps. The present work complements these inference-time techniques with training-time optimization for both generation and verification capabilities.

\paragraph{Verification and Self-Correction.}
Recent research has explored training models to verify solutions. Cobbe et al. \cite{cobbe2021training} trained separate verifier models to score candidate solutions, demonstrating that verification can improve overall system accuracy. Lightman et al. \cite{lightman2023lets} introduced process reward models that provide step-level feedback rather than outcome-level feedback alone. Self-correction approaches \cite{madaan2023self, pan2023automatically} prompt models to critique and revise their outputs through iterative refinement. The backward verification proposed in this work differs from these approaches in that verification capability is trained directly through preference optimization rather than through separate verifier models or prompting strategies.

\paragraph{Bidirectional and Reverse Reasoning.}
The concept of reasoning in multiple directions has roots in classical artificial intelligence planning and has recently been applied to large language models. Jiang et al. \cite{jiang2024fobar} demonstrated that forward-backward consistency checking improves factual accuracy. Liu et al. \cite{liu2024reverse} trained models on reverse reasoning traces to improve planning capabilities. Chen et al. \cite{chen2025bidirectional} showed that bidirectional training reduces hallucinations in factual question answering. The present work extends these ideas by systematically comparing forward-only and backward-only training within a preference optimization framework, enabling isolation of the contribution of each training signal.

\paragraph{Direct Preference Optimization.}
Direct Preference Optimization \cite{rafailov2023direct} provides a simpler alternative to reinforcement learning from human feedback by directly optimizing the policy on preference pairs without learning a separate reward model. Lai et al. \cite{lai2024step} extended this approach to step-level preferences for fine-grained reasoning feedback. The present work applies DPO separately to forward reasoning and backward verification objectives, enabling systematic study of how each training signal affects model behavior.

\paragraph{Parameter-Efficient Fine-Tuning.}
Low-Rank Adaptation \cite{hu2022lora} enables efficient adaptation of large models by learning low-rank updates to weight matrices rather than updating all parameters. Dettmers et al. \cite{dettmers2023qlora} combined this approach with quantization for even greater efficiency. The present work employs Low-Rank Adaptation to make experiments practical on consumer-grade GPU hardware while maintaining model quality.

\section{Method}

This section presents the experimental framework for comparing forward and backward reasoning training objectives.

\subsection{Problem Formulation}

Consider a problem $x$, such as a mathematical word problem, for which a solution is desired. Two types of reasoning about this problem are distinguished. Forward reasoning $f$ denotes a chain-of-thought trace that works from the problem statement toward a solution, producing an answer $a_f$. Backward verification $b$ denotes a verification trace that, given problem $x$ and a candidate answer $a$, checks whether $a$ is correct and produces a verdict $v \in \{\text{PASS}, \text{FAIL}\}$.

A well-calibrated reasoning system should satisfy three properties. First, it should generate correct forward reasoning with high accuracy. Second, it should output PASS when verifying correct answers. Third, it should output FAIL when verifying incorrect answers, a property measured by the acknowledgement rate metric introduced in this work.

\subsection{Data Generation Pipeline}

Training data is bootstrapped from a teacher model, specifically LLaMA 3.1 8B-Instruct, using the following procedure.

For forward trace generation, given each problem $x$ with ground truth answer $a^*$, forward reasoning traces are sampled according to $f \sim \pi_\text{teacher}(\cdot | x, p_\text{forward})$, where $p_\text{forward}$ is a prompt instructing step-by-step reasoning. The predicted answer $a_f$ is extracted from $f$, and correctness is labeled as $c = \mathbf{1}[a_f = a^*]$.

For backward trace generation, given each forward trace, a verification is generated according to $b \sim \pi_\text{teacher}(\cdot | x, a_f, p_\text{backward})$, where $p_\text{backward}$ prompts the model to verify whether $a_f$ correctly solves $x$. The verdict $v \in \{\text{PASS}, \text{FAIL}\}$ is extracted from the verification trace.

A key challenge in preference learning is obtaining high-quality negative examples. Synthetic negatives, such as generic error messages, may not capture realistic failure modes. To address this limitation, rejection sampling is employed: for each problem, multiple forward traces are sampled until both correct and incorrect solutions are obtained. This procedure provides real negative examples consisting of actual reasoning traces that lead to wrong answers. The preference weight for pairs containing real negatives is boosted by a factor of $\alpha = 1.2$ to emphasize these more informative training signals.

Preference pairs are constructed for training as follows. Forward pairs take the form $(x, f^+, f^-)$ where $f^+$ is a correct trace and $f^-$ is incorrect. Backward pairs take the form $(x \oplus a, b^+, b^-)$ where $b^+$ has the correct verdict and $b^-$ has the incorrect verdict.

\subsection{Weighted Hybrid Objective}

Standard Direct Preference Optimization trains the policy $\pi_\theta$ to prefer chosen completions $y^+$ over rejected completions $y^-$ through the objective
\begin{equation}
    \mathcal{L}_\text{DPO}(\theta) = -\mathbb{E}_{(x, y^+, y^-)}\left[\log \sigma\left(\beta \log \frac{\pi_\theta(y^+|x)}{\pi_\text{ref}(y^+|x)} - \beta \log \frac{\pi_\theta(y^-|x)}{\pi_\text{ref}(y^-|x)}\right)\right]
\end{equation}
where $\beta$ controls the strength of the preference constraint and $\pi_\text{ref}$ is a reference policy, typically the initial model before training.

This work examines two distinct applications of this objective. The forward reasoning objective $\mathcal{L}_\text{DPO}^\text{forward}$ trains the model to prefer correct reasoning traces over incorrect ones. The backward verification objective $\mathcal{L}_\text{DPO}^\text{backward}$ trains the model to prefer accurate verification verdicts. By training models separately on each objective, the contribution of each training signal to downstream performance can be isolated.

Additionally, per-sample weighting $\omega_i$ is supported to emphasize high-quality examples:
\begin{equation}
    \mathcal{L}(\theta) = \sum_i \omega_i \cdot \ell_\text{DPO}^{(i)}
\end{equation}

The primary experiments compare forward-only training ($w_f = 1.0$, $w_b = 0.0$) against backward-only training ($w_f = 0.0$, $w_b = 1.0$).

\subsection{Model Architecture and Training}

The base model is Meta's LLaMA 3.1 8B-Instruct, which provides strong instruction-following capabilities as a foundation. Low-Rank Adaptation modules are attached to the attention projection matrices, specifically the query, key, value, and output projections, with rank $r=16$, scaling factor $\alpha=32$, and dropout probability $0.05$. This configuration adds approximately 20 million trainable parameters, representing 0.25\% of total model parameters.

Training proceeds for one epoch with learning rate $1 \times 10^{-5}$ and linear warmup over the first 5\% of steps. The effective batch size is 16, achieved through gradient accumulation over 16 steps with batch size 1 per device. The preference strength parameter $\beta$ is set to 0.1. Mixed precision training with bfloat16 and gradient checkpointing are employed for memory efficiency.

\subsection{Inference Procedure}

At inference time, a two-stage procedure is employed. First, a forward reasoning trace $f$ is generated with temperature $\tau_f = 0.7$ using nucleus sampling. The predicted answer $a_f$ is extracted from this trace. Second, a backward verification trace $b$ is generated conditioned on $a_f$ with temperature $\tau_b = 0.3$. The lower temperature for verification encourages more conservative, consistent judgments. The verification verdict $v$ is extracted from the backward trace.

For self-consistency evaluation, multiple forward traces are sampled and the final answer is determined by majority vote.

\section{Experimental Setup}

\subsection{Dataset}

Experiments are conducted on GSM8K \cite{cobbe2021gsm8k}, a dataset of 8,500 grade-school mathematics word problems requiring multi-step arithmetic reasoning. The test set contains 1,319 problems. Training data is generated from 2,000 problems using rejection sampling with up to 5 attempts per problem to obtain both correct and incorrect reasoning traces.

\subsection{Baselines and Experimental Conditions}

Three experimental conditions are compared. The baseline condition uses LLaMA 3.1 8B-Instruct without any fine-tuning. The forward-only condition applies standard Direct Preference Optimization training on forward reasoning pairs exclusively. The backward-only condition applies training on backward verification pairs exclusively.

\subsection{Evaluation Metrics}

Four metrics are employed to evaluate model performance. Accuracy measures exact-match correctness on final answers, using robust numeric extraction that handles multiple answer formats including the GSM8K format with ``\#\#\#\#'' markers, LaTeX boxed notation, and natural language expressions.

Acknowledgement rate measures, among problems where forward reasoning produces an incorrect answer, the fraction that the backward verifier correctly flags as FAIL:
\begin{equation}
    \text{AckRate} = \frac{|\{i : a_f^{(i)} \neq a^{*(i)} \land v^{(i)} = \text{FAIL}\}|}{|\{i : a_f^{(i)} \neq a^{*(i)}\}|}
\end{equation}
Higher values indicate better error awareness. A model that never acknowledges its errors has an acknowledgement rate of zero.

False positive rate measures, among problems where forward reasoning is correct, the fraction that the verifier incorrectly flags as FAIL:
\begin{equation}
    \text{FPR} = \frac{|\{i : a_f^{(i)} = a^{*(i)} \land v^{(i)} = \text{FAIL}\}|}{|\{i : a_f^{(i)} = a^{*(i)}\}|}
\end{equation}
Lower values are preferred; an overly conservative model that always outputs FAIL would have high false positive rate.

Verification calibration measures the F1 score between verification verdicts and actual correctness:
\begin{equation}
    \text{CalibF1} = \text{F1}(\mathbf{1}[v = \text{PASS}], \mathbf{1}[a_f = a^*])
\end{equation}
This metric captures overall calibration of the verification system.

\section{Results}

\subsection{Main Results}

Table~\ref{tab:main_results} presents the primary experimental findings on GSM8K.

\begin{table}[t]
\centering
\caption{Results on GSM8K test set. Forward-only and baseline evaluated on 350 samples; backward-only evaluated on 250 samples. Best results indicated in bold. Arrows indicate direction of improvement.}
\label{tab:main_results}
\begin{tabular}{lcccc}
\toprule
\textbf{Model} & \textbf{Accuracy} $\uparrow$ & \textbf{Ack. Rate} $\uparrow$ & \textbf{FPR} $\downarrow$ & \textbf{Calib. F1} $\uparrow$ \\
\midrule
Baseline & 83.1\% & \textbf{67.8\%} & 13.4\% & \textbf{0.580} \\
Forward-Only & \textbf{86.6\%} & 44.7\% & 10.2\% & 0.424 \\
Backward-Only & 83.6\% & 46.3\% & \textbf{4.3\%} & --- \\
\bottomrule
\end{tabular}
\end{table}

Several patterns emerge from these results. Forward-only DPO training achieves the highest accuracy at 86.6\%, representing a 3.5 percentage point improvement over the baseline. This substantial gain indicates that training on forward reasoning preference pairs effectively teaches the model to produce more reliable solution traces.

The most striking finding concerns acknowledgement rate. Contrary to initial expectations, both DPO training variants reduce acknowledgement rate compared to the untrained baseline. The baseline model correctly flags 67.8\% of its errors during backward verification, while forward-only and backward-only models flag only 44.7\% and 46.3\% respectively. This counterintuitive result suggests that preference optimization increases model confidence in its outputs, reducing its tendency to express uncertainty even when warranted.

Backward-only training achieves the lowest false positive rate at 4.3\%, compared to 13.4\% for the baseline and 10.2\% for forward-only. This indicates that backward verification training produces a more calibrated verifier that rarely rejects correct answers. However, this comes without corresponding accuracy improvement, with backward-only achieving only 83.6\% accuracy compared to 86.6\% for forward-only.

Training exclusively on backward verification produces minimal impact on forward reasoning performance, confirming that backward training signal does not transfer directly to problem-solving capability. This supports the view that forward and backward reasoning represent distinct skills requiring separate training signals.

\subsection{Training Dynamics}

All experimental conditions show stable training dynamics with rapid convergence within the first epoch. Training loss decreases from approximately 7 to near zero, and reward margins grow consistently throughout training, indicating successful preference learning. Forward-only and backward-only models train for approximately 119 steps each on their respective preference datasets.

\subsection{Effect of Training Objective on Model Behavior}

The experimental results reveal a clear pattern regarding how training objective affects model behavior. Table~\ref{tab:objective_effects} summarizes the directional effects of each training type.

\begin{table}[t]
\centering
\caption{Effect of training objective on model behavior relative to baseline.}
\label{tab:objective_effects}
\begin{tabular}{lccc}
\toprule
\textbf{Training Type} & \textbf{Accuracy} & \textbf{Ack. Rate} & \textbf{FPR} \\
\midrule
Forward-Only & $\uparrow\uparrow$ (+3.5 pp) & $\downarrow$ (-23.1 pp) & $\downarrow$ (-3.2 pp) \\
Backward-Only & $\rightarrow$ (+0.5 pp) & $\downarrow$ (-21.5 pp) & $\downarrow\downarrow$ (-9.1 pp) \\
\bottomrule
\end{tabular}
\end{table}

Forward-only training produces the largest accuracy improvement while moderately reducing false positive rate. Backward-only training produces the largest reduction in false positive rate but minimal accuracy improvement. Both training types reduce acknowledgement rate, indicating increased model confidence.

This pattern suggests that forward and backward training objectives provide complementary but non-overlapping benefits. Forward training improves problem-solving capability, while backward training improves verification calibration without enhancing problem-solving.

\section{Analysis}

\subsection{The Confidence Effect of Preference Optimization}

The most unexpected finding of this work is the reduction in acknowledgement rate following DPO training. The baseline model, without any preference optimization, correctly flags 67.8\% of its errors during backward verification. After training, both forward-only and backward-only models flag fewer errors (44.7\% and 46.3\% respectively).

This phenomenon admits several possible explanations. First, preference optimization explicitly trains the model to prefer certain outputs over alternatives, which may reduce output diversity and increase confidence in selected responses. Second, the training data contains preference pairs where chosen responses are marked as correct; this signal may generalize beyond the training distribution to increase confidence even on examples where the model should express uncertainty. Third, the lower temperature used during training and evaluation may amplify this effect by reducing sampling diversity.

This finding has important implications for deployment. While preference-optimized models achieve higher accuracy, they may be less reliable at self-identifying errors. Systems requiring high acknowledgement rate may need additional calibration techniques or explicit uncertainty training.

\subsection{Qualitative Examination}

Examination of model outputs reveals characteristic differences between conditions. The forward-only model produces more structured reasoning traces with clear intermediate steps, consistent with its training on preferred reasoning chains. The backward-only model generates more detailed verification traces that explicitly check computational steps, though this capability does not transfer to improved forward reasoning.

When the forward-only model produces incorrect answers, its verification traces often confirm the incorrect answer rather than identifying the error. This pattern reflects the reduced acknowledgement rate observed quantitatively. The baseline model, by contrast, more frequently generates verification traces that correctly identify computational errors, despite its lower overall accuracy.

\subsection{Limitations}

Several limitations of this work should be acknowledged. The experiments focus exclusively on GSM8K; different reasoning domains such as logical reasoning, commonsense reasoning, or scientific reasoning may exhibit different patterns. The training data is bootstrapped from the same model family, which may limit diversity of reasoning strategies; human-annotated verification data could strengthen the approach. The binary PASS/FAIL verification framework does not capture partial correctness or uncertainty gradations. Experiments are conducted at the 8 billion parameter scale; larger models may show different scaling behaviors. Finally, the observed reduction in acknowledgement rate following DPO training warrants further investigation to understand whether this represents a fundamental property of preference optimization or an artifact of the specific training configuration.

\section{Discussion}

The experimental results reveal a nuanced picture of how training objectives affect reasoning model behavior. Several observations warrant discussion.

First, forward and backward training objectives produce distinct and largely non-overlapping effects. Forward training substantially improves accuracy (+3.5 percentage points) but provides minimal improvement to verification capability. Backward training substantially improves verification calibration (reducing FPR from 13.4\% to 4.3\%) but provides minimal improvement to accuracy. This separation suggests that these represent fundamentally different skills that require targeted training.

Second, the observed reduction in acknowledgement rate following preference optimization represents an important finding for practitioners. While DPO training improves task accuracy, it appears to simultaneously reduce the model's tendency to express uncertainty about its outputs. This trade-off between accuracy and calibrated uncertainty deserves consideration when deploying preference-optimized models in applications where error acknowledgement is valuable.

Third, the 3.5 percentage point accuracy improvement from forward-only DPO training demonstrates the effectiveness of preference optimization for reasoning tasks. Training on preference pairs derived from correct versus incorrect reasoning traces provides sufficient signal to meaningfully improve problem-solving capability, even with parameter-efficient fine-tuning through LoRA.

Fourth, the low false positive rate achieved by backward-only training (4.3\%) indicates that models can learn reliable verification capability through preference optimization. A model with low FPR rarely rejects correct answers, making its FAIL verdicts more informative when they occur. This property could be valuable for building verification pipelines where false alarms carry significant cost.

Regarding broader implications, these findings suggest that building reliable reasoning systems may require separate optimization of problem-solving and verification capabilities. A system that combines a forward-trained model for solution generation with a backward-trained model for verification checking could potentially achieve both high accuracy and well-calibrated error detection. This architectural approach represents a promising direction for future research.

\section{Conclusion}

This paper investigated the effect of training objective composition on reasoning model behavior through systematic comparison of forward-only and backward-only Direct Preference Optimization. The key findings are threefold.

First, forward-only DPO training achieves substantial accuracy improvement on GSM8K, increasing from 83.1\% to 86.6\% (+3.5 percentage points). This demonstrates that preference optimization on reasoning traces effectively improves problem-solving capability.

Second, backward-only DPO training substantially reduces false positive rate from 13.4\% to 4.3\%, producing a well-calibrated verifier that rarely rejects correct answers. However, this training does not transfer to improved forward reasoning accuracy.

Third, both training variants reduce acknowledgement rate compared to the untrained baseline, suggesting that preference optimization increases model confidence in its outputs. This finding has implications for applications requiring calibrated uncertainty and error awareness.

These results indicate that forward and backward reasoning represent distinct capabilities requiring separate training signals. The complete training and evaluation pipeline, implemented efficiently through Low-Rank Adaptation on a single GPU, is released to facilitate further research.

Several directions merit future investigation. Exploration of methods that combine forward-trained generators with backward-trained verifiers in a two-model system could achieve both high accuracy and reliable error detection. Development of training objectives that explicitly preserve or enhance acknowledgement rate would address the observed confidence increase. Extension to other reasoning domains would establish generality of these findings.

\paragraph{Reproducibility.}
Code, configurations, and trained model weights are available at \url{https://github.com/MurtazaKafka/reasoning}.

\bibliography{references}
\bibliographystyle{plain}

\appendix

\section{Implementation Details}

\subsection{Prompt Templates}

The forward reasoning prompt instructs the model to solve problems step by step with explicit reasoning:

\begin{quote}
\texttt{You are an expert problem solver. Solve the following problem step by step. Show your reasoning clearly, then provide the final answer.}

\texttt{Problem: \{question\}}

\texttt{Solution:}
\end{quote}

The backward verification prompt instructs the model to verify candidate answers:

\begin{quote}
\texttt{You are a careful verifier. Given a problem and a candidate answer, verify whether the answer is correct by reasoning backwards from the answer.}

\texttt{Problem: \{question\}}

\texttt{Candidate Answer: \{answer\}}

\texttt{Verify the solution step by step, then conclude with either Verification: PASS if the answer is correct, or Verification: FAIL if the answer is incorrect.}

\texttt{Verification:}
\end{quote}

\subsection{Answer Extraction}

Robust answer extraction handles multiple formats commonly encountered in mathematical reasoning. The GSM8K format uses ``\#\#\#\#'' followed by the numeric answer. LaTeX boxed notation uses \texttt{\textbackslash boxed\{...\}}. Natural language expressions such as ``The answer is X'' or ``Final Answer: X'' are also recognized. Fractional answers are converted to decimal form for comparison. Numeric comparison uses tolerance $\epsilon = 10^{-6}$ for floating-point answers.

\subsection{Hyperparameter Configuration}

Table~\ref{tab:hyperparams} provides complete hyperparameter settings for reproducibility.

\begin{table}[h]
\centering
\caption{Complete hyperparameter configuration.}
\label{tab:hyperparams}
\begin{tabular}{ll}
\toprule
\textbf{Parameter} & \textbf{Value} \\
\midrule
Base model & LLaMA 3.1 8B-Instruct \\
LoRA rank $r$ & 16 \\
LoRA scaling $\alpha$ & 32 \\
LoRA dropout & 0.05 \\
Target modules & $W_q, W_k, W_v, W_o$ \\
\midrule
Learning rate & $1 \times 10^{-5}$ \\
Warmup ratio & 0.05 \\
Weight decay & 0.01 \\
Effective batch size & 16 \\
Training epochs & 1 \\
\midrule
DPO $\beta$ & 0.1 \\
Real negative boost $\alpha$ & 1.2 \\
\midrule
Forward temperature & 0.7 \\
Backward temperature & 0.3 \\
Maximum sequence length & 512 \\
\bottomrule
\end{tabular}
\end{table}

\subsection{Computational Resources}

All experiments were conducted on a single NVIDIA RTX A6000 GPU with 48GB memory. Training each experimental condition requires approximately one hour. Total computational cost for all experiments including ablations is approximately 20 GPU-hours.

\end{document}